\documentclass{ifacconf}

\usepackage{graphicx}      % include this line if your document contains figures
\usepackage{natbib}        % required for bibliography
\usepackage{algorithm,algpseudocode}
\usepackage{amsfonts}
\usepackage{float}
\usepackage{comment}
\usepackage{amsmath}
\begin{document}
\begin{frontmatter}

\title{Cascade Attribute Network: Decomposing Reinforcement Learning Control Policies using Hierarchical Neural Networks} 

\thanks[footnoteinfo]{The first two authors contributed equally to the work.}

\author[First]{Haonan Chang} 
\author[Second]{Zhuo Xu} 
\author[Second]{Masayoshi Tomizuka} 

\address[First]{Dept. of Robotics, University of Michigan, Ann Arbor. (e-mail: harveych@umich.edu).}
\address[Second]{Dept. of Mechanical Engineering, University of California, Berkeley. (e-mail: \{zhuoxu,tomizuka\}@berkeley.edu)}

\begin{abstract}                % Abstract of not more than 250 words.
Reinforcement learning methods have been developed to achieve great success in training control policies in various automation tasks. However,  a main challenge of the wider application of reinforcement learning in practical automation is that the training process is hard and the pretrained policy networks are hardly reusable in other similar cases. To address this problem, we propose the cascade attribute network (CAN), which utilizes its hierarchical structure to decompose a complicated control policy in terms of the requirement constraints, which we call attributes, encoded in the control tasks. We validated the effectiveness of our proposed method on two robot control scenarios with various add-on attributes. For some control tasks with more than one add-on attribute attributes, by directly assembling the attribute modules in cascade, the CAN can provide ideal control policies in a zero-shot manner.
\end{abstract}

\begin{keyword}
Reinforcement learning control,Deep Learning, Artificial Intelligence
\end{keyword}

\end{frontmatter}
%===============================================================================

\section{INTRODUCTION}

Reinforcement learning (RL) is an artificial intelligence approach that solves for automatic control policies, $\pi(a|s)$, that map the state space inputs $s$ to the control command outputs $a$. The training of the RL policies is through the interaction of the agent and the environment, which is often modeled using a Markov Decision Process (MDP). The RL has been successful in solving many robotics and automation problems in simulations as well as real-world scenarios (\cite{visuomotor})(\cite{rainbow})(\cite{navigation}). However, the wide application of the RL in automation is slowed down by several challenges, and one of the main challenges is the difficulty of RL policy training. The RL policy generally takes a great amount of computation power to train for complicated control tasks, and what makes it worse is that the RL policies are optimized based on certain fixed MDPs, and the knowledge encoded inside an optimized control policy are hard to transfer to other similar tasks. That is, new RL policies have to be trained from scratch even for those control tasks that are very similar to the pretrained task, which leads to a great amount of computation power waste. 

Unlike the existing transfer learning approaches (\cite{progressive1}) (\cite{transferable_policy}) that are lack of interpretable transferable features, we propose to decompose the complicated control problems in terms of its consisting requirement constraints, which we call attributes. The conception attributes refer especially to global characteristics or requirements that take effect throughout the task. For example, to solve an autonomous driving problem, we can first decompose the requirements of the task into a base target reaching attribute, an add-on obstacle avoidance and a speed limit attribute. Our methodology includes training an attribute module for each of the attributes and then assembling the attribute modules together to produce the overall policy. To deal with tasks under different attributes, we propose an RL framework called the cascade attribute network (CAN). In CAN, the attribute modules are connected in the cascade series. Our method has two main intriguing advantages:

\begin{enumerate}
\item The decomposed attribute modules are much easier and faster to train compared with the overall control policy.
\item The attribute related knowledge is encoded in interpretable modules, which can be built up to create versatile policies that can adjust to changes in the control tasks.
\end{enumerate}

% The remainder of this paper is organized as follows: the related works and the background of RL are introduced in Section II. In Section III, the architecture of the CAN and the implementation details are described. In Section IV, we show simulation results to validate the proposed model using a variety of robots and attributes. The conclusions are given in Section V.

\section{Related Work and Background}

\subsection{Related Work}

There have been lots of attempts to create versatile intelligence that can adjust to changes in the tasks as well. Transfer learning (\cite{reinforcement_transfer}) is a key tool that makes the use of previously learned knowledge for the better or faster learning of new knowledge. Rusu et al (\cite{progressive1})(\cite{progressive2}) establish a progressive network, in which newly added columns are laterally connected to previously learned columns for knowledge transfer. Drafty et al (\cite{transferable_policy}) and Braylan et al (\cite{reuse_module}) also design network architectures for knowledge transfer in MAV control and video game playing. Barreto et al (\cite{successor1})(\cite{successor2}) propose a scheme based on successor features, a value function describing the dynamics of the environment and a generalized policy improvement over multiple policies. In this way, the method can provide an efficient transferring among different tasks where the reward function is changed but the dynamics of the system remains the same. As for the combinations of transfer learning and imitation learning, Ammar et al (\cite{unsupervised_transfer}) use unsupervised learning to map states for transfer, assuming the existence of distance function between different state spaces. Gupta et al (\cite{invariant_feature}) learn an invariant feature between different dimensional states and use demonstrations to increase the density of the rewards. Our recent works (\cite{xu2018zero})(\cite{tang2019disturbance}) on transfer learning utilizes interpretable trajectory as transferrable feature, but the human engineered features lack the versatility to be applied in various robotic control areas.
% Our work differs from those previous works mainly in that we put emphasis on the modularization of attributes, which are concrete and meaningful modules that can be conveniently assembled into various combinations.

There are other methods seeking to learn a globally general policy:
Curriculum learning (CL)(\cite{curriculum}) trains a model on a sequence of cognate tasks that get more and more challenging gradually, so as to solve hard tasks that could not be learned from scratch. Florensa et al (\cite{reverse_curriculum}) apply reverse curriculum generation (RCL) in RL. In the early stage of the training process, the RCL initializes the agent state to be very close to the target state, making the policy very easy to train. They then gradually increase the random level of the initial state as the RL model performs better and better. Sanmit et al (\cite{CLMDP}) formulate the curriculum learning sequence as an MDP process, which can also be learned from experience. These results show that with a good consideration of the nature of the problem, CL can help reinforcement learning have a better convergence speed. Our policy training strategy is inspired by the idea of CL and achieved satisfying robustness for the policies. There are also researches in training modular neural networks, (\cite{modular_robot_task}) investigates the combinations of multiple robots and tasks, while (\cite{modular_subtask}) investigates the combinations of multiple sequential subtasks. Our work, along with our parallel work (\cite{xu2019toward}) different from those works, looks into modularization in a different dimension. We investigate the modularization of attributes, the characteristics or requirements that take effect throughout the whole task.

\section{The Cascade Attribute Networks}

\subsection{Problem Formulation}
We consider an agent with a state-space $A$ performing a class of tasks built up with one base attribute and a series of add-on attributes. We model the control tasks and the attribute in a unified form using MDPs.  We denote the attributes using their index $\{ 0,1,2,\ldots \}$, where the $0^{th}$ attribute is the base attribute, which usually corresponds to the most fundamental goal of the task, such as the target reaching attribute in the autonomous driving task.  We define the state space of each attribute to be the minimum state space that fully characterizes the attribute, denoted $S=\{S_0, S_1, S_2, S_3\ldots \}$. For example, let the base attribute be the target reaching attribute, and the $1^{st}$ attribute be the obstacle avoidance attribute. Then $S_0$ consists of the states of the agent and the target, while $S_1$ consists of the states of the agent and the obstacle, but does not include the states of the target. Using the attribute as elements, one can build various control tasks, such as pure target reaching, target reaching while avoiding an obstacle, target reaching under external force influence, and so on. 

Each attribute has an unique reward function as well, denoted $R=\{R_0,R_1,R_2,R_3\ldots\}$. Each $R_i$ is a function mapping a state action pair to a real number reward, i.e. $R_i: S_i \times A \rightarrow \mathbb{R}$. 

\subsection{Network Architecture}
\begin{figure}[t]
\centering
\includegraphics[scale=0.35]{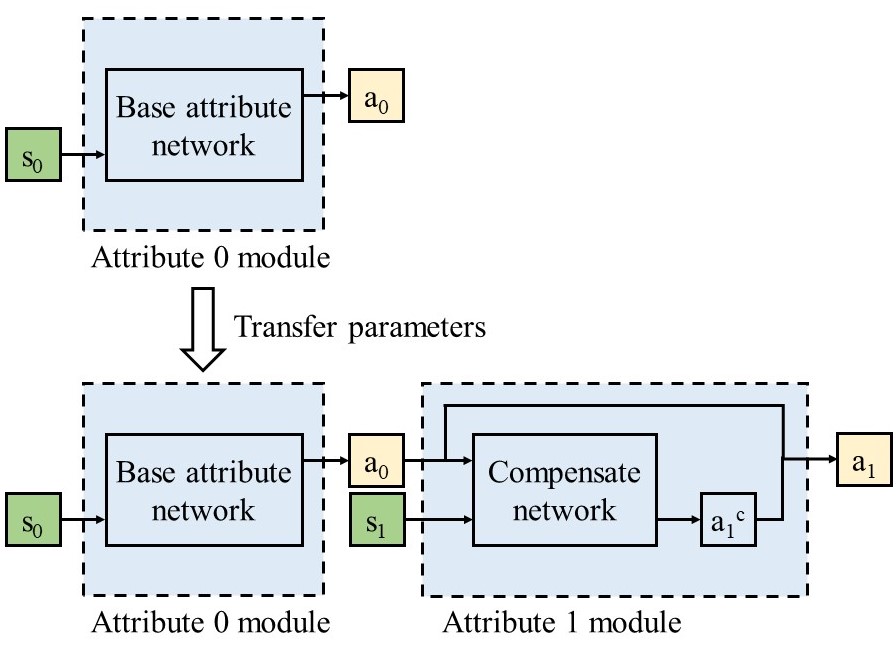}
\label{figurelabel}
\caption{The training procedure of an attribute module in the CAN: first train the base attribute module, then train the add-on attribute module based on fixed pretrained base module}
\label{sa}
\end{figure}

The architecture of the CAN is shown in Fig. \ref{sa}. The training for a task with more than one attribute is divided into two parts: the training of the base attribute module and the add-on attribute module. In the training phase, first, an RL policy $\pi_0(a_0|s_0)$ is trained to accomplish the goal of the base attribute. The base attribute network takes in $s_0\in S_0$ and outputs $a_0 \in A$, the reward function of the MDP is given by $R_0$. This process is a default RL training process. 

Consider the control task with more than one attribute, without loss of generality, the $1^{st}$ add-on attribute module is trained next, in a series of the base attribute module. The $1^{st}$ attribute module consists of a compensation network and a weighted sum operator. The compensation network is fed with state $s_1 \in S_1$, and action $a_0$ chosen by $\pi_0$. The output of the compensation network is the compensation action $a^{c}_1$, which is used to compensate $a_0$ to produce the overall action $a_1$. The reward for the MDP is given by $R_0+R_1$ so that the requirements for both constraints of attributes are satisfied. Since the parameters of the base attribute network are pretrained and fixed, the cascading attribute network would extract the features of the attribute by exploring the new MDP under the guidance of the base policy. It is noted that in the add-on attribute module, the weight of the compensation action $a^{c}_1$ is initiated to be small and increased over the training time. That is, at the early stage of the training process, mainly $a_0$ takes effect, while $a^{c}_1$ gradually gets to influence the overall $a_1$ as the training goes on. For the other attributes, the training of their attribute networks is the same. 

\begin{figure}[t]
\centering
\includegraphics[scale=0.38]{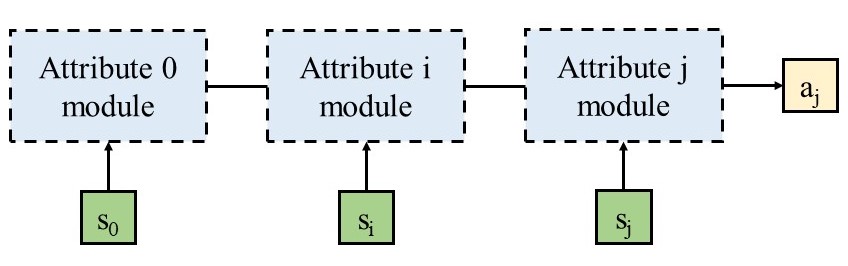}
\label{figurelabel}
\caption{One kind of usage of the CAN in multi-add-on-attributes tasks, by assembling add-on attribute modules in cascade to the base attribute module, the output action of the last attribute module is the outcome of the overall hierarchical policy network.}
\label{ma}
\end{figure}

Empirical results also show that for cases, where the number of the add-on attributes is small, and different add-on attributes do not entangle with each other, with carefully designed training schemes, the CAN shown in Fig. \ref{ma} can be used to provide an ideal control policy. In the CAN shown in Fig. \ref{ma}, the $i^{th}$ attribute module takes in $s_i$ and $a_{i-1}$, and outputs $a_i$ that satisfies all the attribute before the $i^{th}$ module. The final output $a_j$ is the overall output that satisfies all the constraints in the constraint array.

\section{Experiments}

\subsection{Environmental Setup}
The experiments are powered by the MuJoCo physics simulator (\cite{mujoco}). The attribute modules in our experiments are all three-layer fully connected networks which output Gaussian distributed stochastic actions, built using TensorFlow. The baseline RL algorithm we use is the PPO (\cite{ppo}) method with GAE (\cite{gae}) as the advantage estimator. 

We evaluate the capability of the CAN on two types of robot scenarios in our experiments. One is a point robot with a 2-dimensional action space, and the other is an articulated robot with a 5-dimensional action space. For the point robot, the state space includes the position and velocity of the robot, and the action vector is the driving force applied to it. For the articulated robot the state spaces are the angle and angular velocity at all the joints, and the x-position and speed of the base of the robot, while the action vector includes the torques at all the joints, and the force applied to the base of the robot. For each robot agent, we implement one base attribute and four different add-on attributes. For each different attribute, we designed a reward function $R_i$. The attribute settings and reward functions are described as follow:

\begin{figure}[t]
\centering
\includegraphics[scale=0.35]{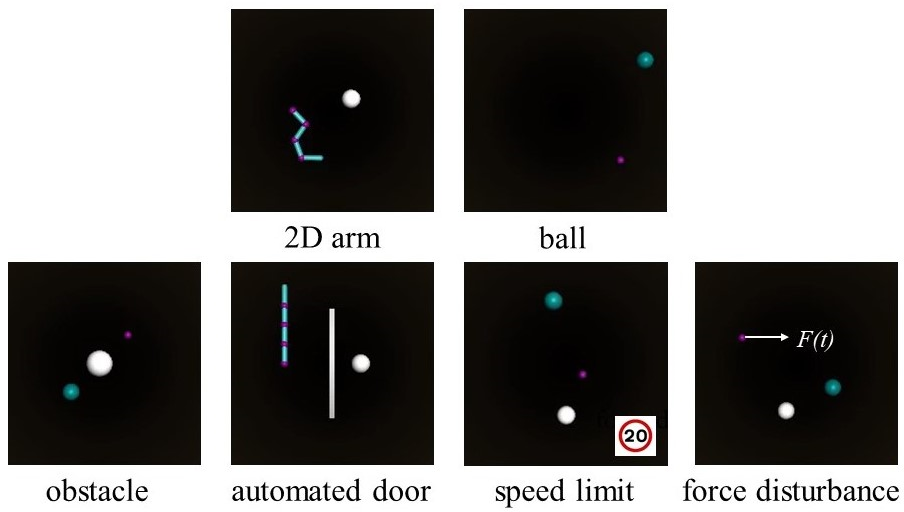}
\label{figurelabel}
\caption{The images in the top row show the two robot scenarios, in which the agents are performing the base attribute of target reaching. The bottom images show the four add-on attributes.}
\end{figure}

\subsubsection{Reaching (base attribute)}
Target reaching task, a common robot task, is defined to be the base attribute for both robot scenarios in our experiment. Given initial configuration and location of the target, the robots aim to to reach the target zone using their end effector. The reward function is shown below:
\[
r_b(t)=\left\{
\begin{aligned}
1, &\ reaches\ target \\ 
0, &\ other\ case
\end{aligned}
\right.\eqno{(4)}
\]

\subsubsection{Obstacle (position phase)}
In our obstacle avoidance attribute, a moving circular obstacle is placed in the work area. In our implementation, one obstacle avoidance attribute correspond to one circular obstacle. Nevertheless, many obstacle attributes are allowed to be assembled together to create a more complicated environment. The reward function is represented as following:
\[
r_o^i(t)=\left\{
\begin{aligned}
-0.3, &\ touches\ i^{th} obstacle \\ 
0, &\ other\ case
\end{aligned}
\right.\eqno{(5)}
\]
\subsubsection{Automated door (time phase)}
The automated door attribute is a time controlled obstacle. This door is closed in the first and will open at some specific time slots. This attribute is challenging to train using RL since it punishes the agent even if it goes to the right direction at a wrong time.  The reward function is designed as following:

\[
r_d(t)=\left\{
\begin{aligned}
-0.01, &\ touches\ door \\ 
0, &\ other\ case
\end{aligned}
\right. \eqno{(6)}
\]
\subsubsection{Speed limit (velocity phase)}
Practical robots always have dynamic constraints on their joints which limit their speeds. The speed limit at time t, L(t), is designed to be a time-variant function as shown in Fig \ref{fas}. Denote the maximum joint velocity as $v_{max}$. Reward function can be written as:

\[
r_s(t)= -0.3(max(v_{max} - L(t),0))\eqno{(7)}
\]

\subsubsection{Force disturbance (acceleration phase)}
In working scenarios, a robot can be influenced by force disturbance or repulsion force on their joints. The force disturbance attribute in our experiment corresponds to a time-invariant force disturbance added to a certain joint of both point robot and articulated robot. 

The force disturbance only affects the dynamics function of the robot system, with no additional reward function added.
\[
r_f(t)=0 \eqno{(8)}
\]

\subsection{Training Schemes}
To guarantee the capacity of the CAN, the attribute modules need to meet two requirements:

\begin{enumerate}
\item The base attribute policies should be robust over the state space, rather than being effective only at the states that are close to the optimal trajectory. This enables the base attribute policies to be instructive when a compensation action is added on the top of it.
\item The compensation action for a certain attribute should be close to zero if the agent is in a state where this attribute is not active. This property increases the capability of multi-attribute structures.
\end{enumerate}

For the sake of the robustness of the attribute policies, we apply CL to learn a general policy that can accomplish the task starting from any initial state. The CL algorithm first trains a policy with a fixed initial state. As the training goes on, the random level of the initial state is smoothly increased, until the initial state is randomly sampled from the whole state space. The random level is increased only if the policy is capable enough for the current random level, which is reflected by the increase of the episodic reward. The pseudocode for this process is shown in Algorithm 1. 

\begin{algorithm}[t]
\caption{Curriculum Learning}
\label{RCL}
\begin{algorithmic}[1]
\State $RandomLevel = $ Initial Random Level
\State $\lambda = 1 + $ Random Level Increase Rate
\State $N = $ Batch Number
\State $LongTermR = Queue()$
\While{$RandomLevel < \textrm{Terminal Random Level}$}
    \State Update the policy using PPO
    \State $Rewards \gets RunEpisode(N)$ 
    \State $LongTermR.append(Rewards)$
    \If{$Average(LongTermR) > Threshold$}
       \State $RandomLevel = RandomLevel \times \lambda$
       \State $Clear(LongTermR)$
    \EndIf
\EndWhile
\end{algorithmic}
\end{algorithm}

To guarantee the second requirement, an extra loss term that punishes the magnitude of the compensative action, $l^{c}_i \propto -\|a^{c}_i\|^2$, is added to the reward function so as to reduce $\|a^{c}_i\|$ when attribute $i$ is not active.

\subsection{CAN Performance}
The first set of experiments test the capability of the CAN to learn attributes and assemble learned attributes. We first train the base attribute module using the baseline RL algorithm with CL and then use the cascading modules to decompose the different add-on attributes based on the pretrained base module. In the actor-critic RL training using PPO, the maximum episodes is 10000. Both the actor network and critic network are trained using the Adam optimizer, with a batch number of 256, an initial learning rate of 0.0001, and are updated for 20 times in each training iteration. After training the two robot scenarios and all the four add-on attributes for each scenario, tasks. Therefore, the results show that the add-on attributes can be successfully added to the base attribute using the CAN. Fig. 4. shows the random level training log for the four add-on attribute modules in the point mass scenario,  we ran 10 test episodes for the 8 combination tasks and received 10/10 success rate for all the tasks and Fig. 5 and Fig. 6 show the example episodes of the performing different attributes combinations.

We also show cases where the CAN can assemble more than one add-on attribute modules. Concretely, we first train the obstacle attribute module based on the point mass robot. Then we connect two identically parameterized obstacle attribute modules in series following the base attribute module, each handling one obstacle ball. The CAN structure is the same as the one shown in Fig. 2.. Fig. 7. shows two examples of the CAN zero shooting the task where the moving point robot reaches the target while avoiding two obstacles simultaneously. Our perspective is not learning a universally effective policy for any control task, but utilizing the decomposition of the policy to reuse pre-learned knowledge, so as to achieve easier and faster training compared to training a new policy from scratch for new control task (as shown in Section IV-C).

\begin{figure}[t]
    \centering
    \includegraphics[width=0.9\linewidth]{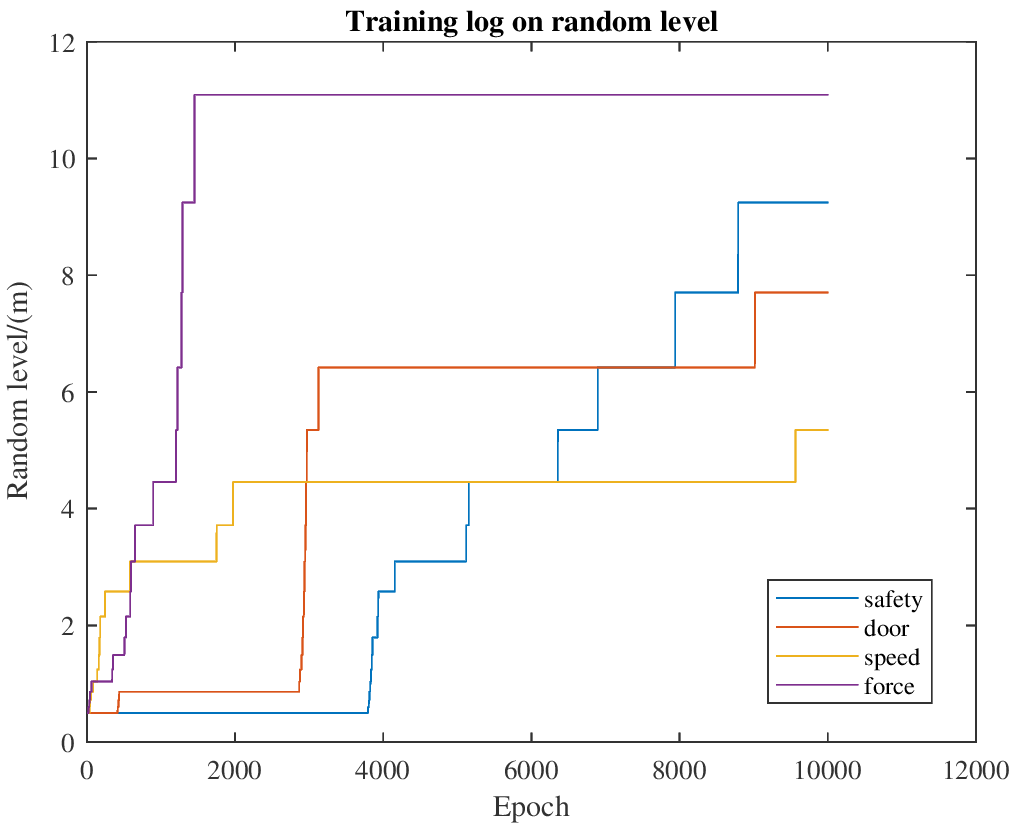}
    \caption{Training log on random level for the point mass robot with the four add-on attributes. CL is applied, and the improvement of the random level is based on that the total reward performance being consistently improving.}
    \label{random_log}
\end{figure}

\begin{figure*}[t]
\centering
\includegraphics[scale=0.45]{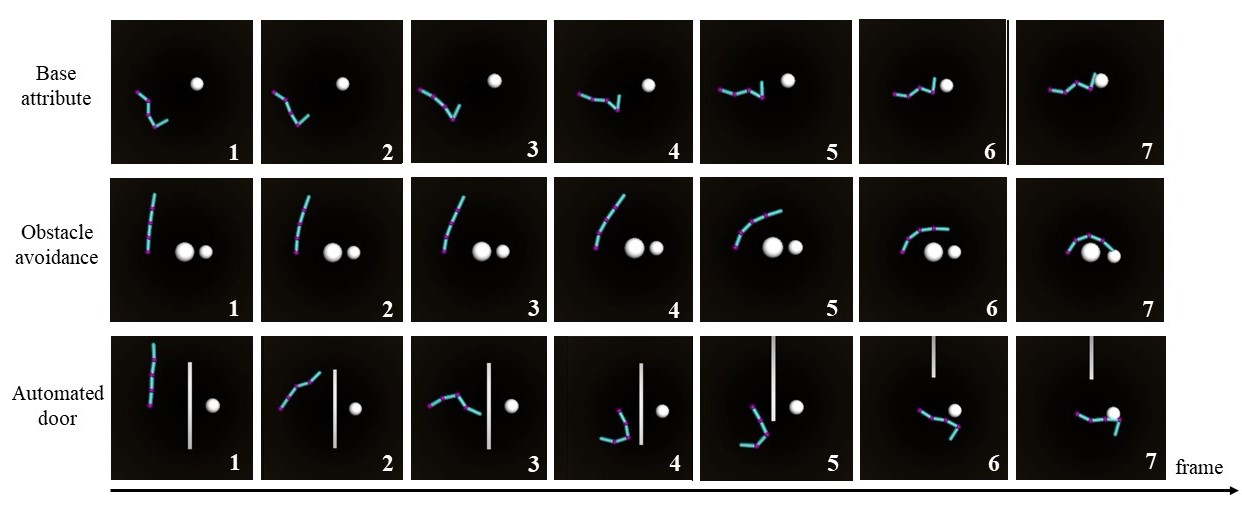}
\label{figurelabel}
\caption{Three example episodes for the articulated robot: target reaching, target reaching while avoiding an obstacle, or an automated door.}
\end{figure*}
\begin{figure*}[t]
    \centering
    \hbox{\hspace{5em} \includegraphics[scale=0.41]{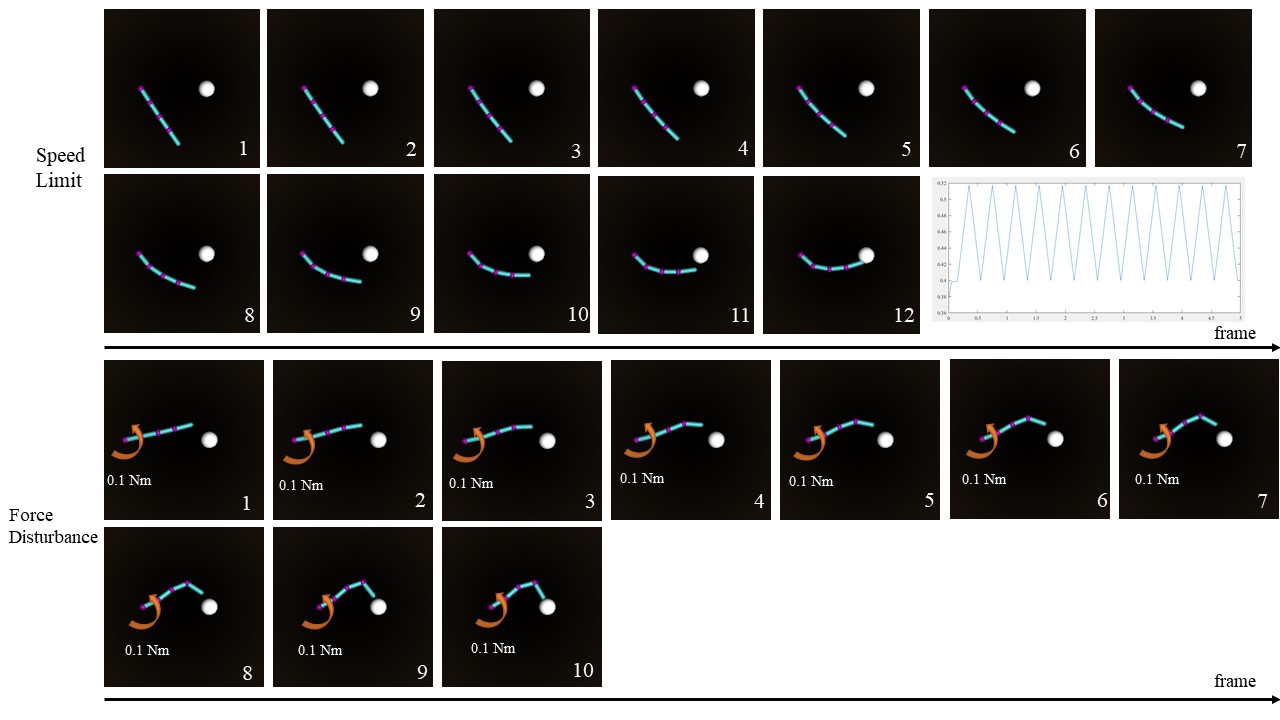}}
    \caption{Two example episodes for the articulated robot: target reaching under the speed limit, or force disturbance. Because of the limit of speed and force disturbance in the actuator, these two tasks take a longer time to finish compared to the three previous tasks. The speed limit and external force are also visualized.}
    \label{fas}
\end{figure*}

\begin{figure*}[t]
\centering
\includegraphics[scale=0.45]{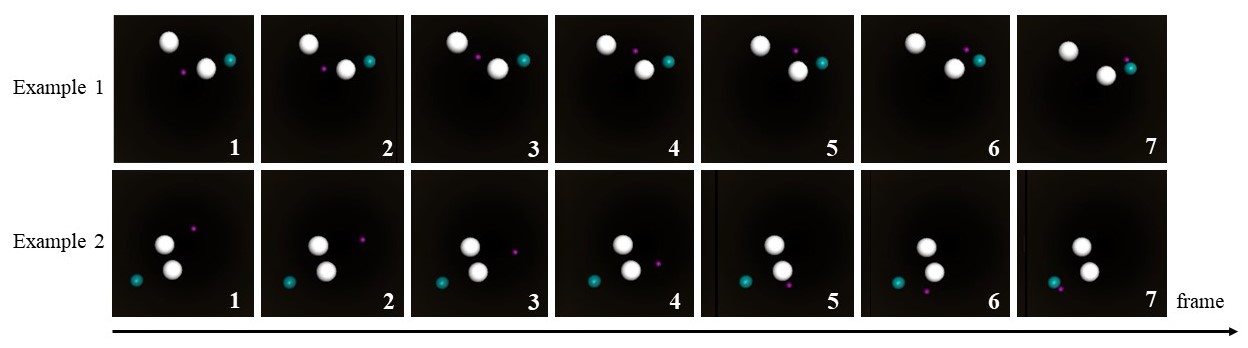}
\label{figurelabel}
\caption{Moving point robot (the pink ball) reaches the target (the green ball) while avoiding two obstacles simultaneously. Note that this task is never seen by the CAN. The obstacle attribute module is trained only once, and two identically parameterized obstacle attribute modules corresponding two different obstacles are assembled together to zero shoots the new task.}
\end{figure*}

\subsection{Comparison with Baseline RL Methods}
We compare the capability and efficiency of the CAN and the baseline RL on an obstacle task for the point mass robot. The CAN is trained with CL, with the base attribute pretrained and fixed. For the baseline RL, the task is trained from scratch. For baseline RL trained with RCL, we let the initial state be very close to the target in the early stage of the training phase. Using RCL, the baseline RL could gain positive reward in the early stage. The challenge would be whether the RL algorithm can maintain a high reward level as the random level increases.

The resulting comparison of the reward and random level are shown in Fig. 8, with the maximum training iterations exceeding 120000. It is shown that the baseline RL using CL barely learns anything. This is because the reward is too sparse and the agent is consistently receiving punishment from the obstacle and falls into a local minimum of purely avoiding the obstacle and ignoring the target. For the baseline RL using RCL, in the early stage, the average discounted reward in an episode is high as expected. But as the random level rises, the performance of the baseline RL with RCL drops. Therefore, the random level increases slowly as the training goes on. The CAN, on the other hand, is able to overcome the misleading punishments from the obstacle. As a result, the random level of the CAN rises rapidly, and the CAN achieves terminal random level more than 10 times faster than the baseline. 

\begin{figure}[t]
\centering
\includegraphics[scale=0.5]{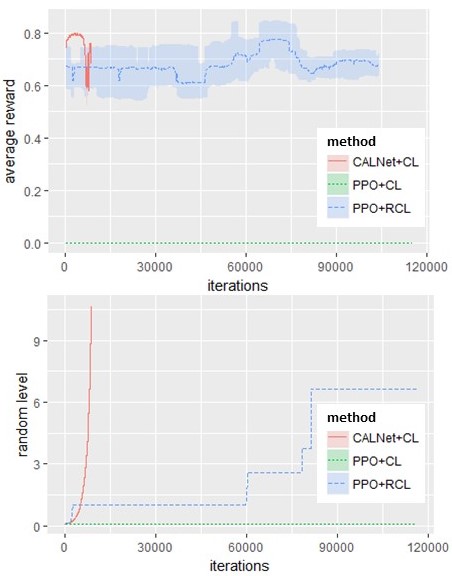}
\label{comparision}
\caption{Comparison between the performance of the CAN and the baseline RL (PPO) in the training phase.}
\end{figure}

\section{Conclusions and Future Work}

In this paper, we propose the attribute learning and present the advantages of using this novel method to decompose the control policies of complicated control tasks. The RL framework we propose, the CAN, uses cascading attribute modules structure to model the characteristics of the attributes. The attribute modules are trained with the guidance of the pretrained base attribute module. We validate the effectiveness of the CAN of decomposing and assembling attributes and show the advantages of the CAN in solving complicated tasks compared to the baseline RL. Our ongoing future work includes applying the CAN method on more advanced simulation scenarios such as autonomous driving and using a real 2D articulated robot similar to the one in the simulation experiment to verify the capability of the CAN in more general and complicated settings.

\bibliography{ifacconf}   

\begin{thebibliography}{23}
\providecommand{\natexlab}[1]{#1}
\providecommand{\url}[1]{\texttt{#1}}
\providecommand{\urlprefix}{URL }
\expandafter\ifx\csname urlstyle\endcsname\relax
  \providecommand{\doi}[1]{doi:\discretionary{}{}{}#1}\else
  \providecommand{\doi}{doi:\discretionary{}{}{}\begingroup
  \urlstyle{rm}\Url}\fi

\bibitem[{Ammar et~al.(2015)Ammar, Eaton, Ruvolo, and
  Taylor}]{unsupervised_transfer}
Ammar, H.B., Eaton, E., Ruvolo, P., and Taylor, M.E. (2015).
\newblock Unsupervised cross-domain transfer in policy gradient reinforcement
  learning via manifold alignment.
\newblock In \emph{Twenty-Ninth AAAI Conference on Artificial Intelligence}.

\bibitem[{Andreas et~al.(2017)Andreas, Klein, and Levine}]{modular_subtask}
Andreas, J., Klein, D., and Levine, S. (2017).
\newblock Modular multitask reinforcement learning with policy sketches.
\newblock In \emph{Proceedings of the 34th International Conference on Machine
  Learning-Volume 70}, 166--175. JMLR. org.

\bibitem[{Barreto et~al.(2019)Barreto, Borsa, Quan, Schaul, Silver, Hessel,
  Mankowitz, {\v{Z}}{\'\i}dek, and Munos}]{successor2}
Barreto, A., Borsa, D., Quan, J., Schaul, T., Silver, D., Hessel, M.,
  Mankowitz, D., {\v{Z}}{\'\i}dek, A., and Munos, R. (2019).
\newblock Transfer in deep reinforcement learning using successor features and
  generalised policy improvement.
\newblock \emph{arXiv preprint arXiv:1901.10964}.

\bibitem[{Barreto et~al.(2017)Barreto, Dabney, Munos, Hunt, Schaul, van
  Hasselt, and Silver}]{successor1}
Barreto, A., Dabney, W., Munos, R., Hunt, J.J., Schaul, T., van Hasselt, H.P.,
  and Silver, D. (2017).
\newblock Successor features for transfer in reinforcement learning.
\newblock In \emph{Advances in neural information processing systems},
  4055--4065.

\bibitem[{Bengio et~al.(2009)Bengio, Louradour, Collobert, and
  Weston}]{curriculum}
Bengio, Y., Louradour, J., Collobert, R., and Weston, J. (2009).
\newblock Curriculum learning.
\newblock In \emph{Proceedings of the 26th annual international conference on
  machine learning}, 41--48. ACM.

\bibitem[{Braylan et~al.(2016)Braylan, Hollenbeck, Meyerson, and
  Miikkulainen}]{reuse_module}
Braylan, A., Hollenbeck, M., Meyerson, E., and Miikkulainen, R. (2016).
\newblock Reuse of neural modules for general video game playing.
\newblock In \emph{Thirtieth AAAI Conference on Artificial Intelligence}.

\bibitem[{Daftry et~al.(2016)Daftry, Bagnell, and Hebert}]{transferable_policy}
Daftry, S., Bagnell, J.A., and Hebert, M. (2016).
\newblock Learning transferable policies for monocular reactive mav control.
\newblock In \emph{International Symposium on Experimental Robotics}, 3--11.
  Springer.

\bibitem[{Devin et~al.(2017)Devin, Gupta, Darrell, Abbeel, and
  Levine}]{modular_robot_task}
Devin, C., Gupta, A., Darrell, T., Abbeel, P., and Levine, S. (2017).
\newblock Learning modular neural network policies for multi-task and
  multi-robot transfer.
\newblock In \emph{2017 IEEE International Conference on Robotics and
  Automation (ICRA)}, 2169--2176. IEEE.

\bibitem[{Florensa et~al.(2017)Florensa, Held, Wulfmeier, Zhang, and
  Abbeel}]{reverse_curriculum}
Florensa, C., Held, D., Wulfmeier, M., Zhang, M., and Abbeel, P. (2017).
\newblock Reverse curriculum generation for reinforcement learning.
\newblock \emph{arXiv preprint arXiv:1707.05300}.

\bibitem[{Gupta et~al.(2017)Gupta, Devin, Liu, Abbeel, and
  Levine}]{invariant_feature}
Gupta, A., Devin, C., Liu, Y., Abbeel, P., and Levine, S. (2017).
\newblock Learning invariant feature spaces to transfer skills with
  reinforcement learning.
\newblock \emph{arXiv preprint arXiv:1703.02949}.

\bibitem[{Hessel et~al.(2018)Hessel, Modayil, Van~Hasselt, Schaul, Ostrovski,
  Dabney, Horgan, Piot, Azar, and Silver}]{rainbow}
Hessel, M., Modayil, J., Van~Hasselt, H., Schaul, T., Ostrovski, G., Dabney,
  W., Horgan, D., Piot, B., Azar, M., and Silver, D. (2018).
\newblock Rainbow: Combining improvements in deep reinforcement learning.
\newblock In \emph{Thirty-Second AAAI Conference on Artificial Intelligence}.

\bibitem[{Kahn et~al.(2018)Kahn, Villaflor, Ding, Abbeel, and
  Levine}]{navigation}
Kahn, G., Villaflor, A., Ding, B., Abbeel, P., and Levine, S. (2018).
\newblock Self-supervised deep reinforcement learning with generalized
  computation graphs for robot navigation.
\newblock In \emph{2018 IEEE International Conference on Robotics and
  Automation (ICRA)}, 1--8. IEEE.

\bibitem[{Levine(2016)}]{visuomotor}
Levine, S. (2016).
\newblock End-to-end training of deep visuomotor policies.
\newblock \emph{Journal of Machine Learning Research.}, 17.39, 1--40.

\bibitem[{Narvekar and Stone(2019)}]{CLMDP}
Narvekar, S. and Stone, P. (2019).
\newblock Learning curriculum policies for reinforcement learning.
\newblock In \emph{Proceedings of the 18th International Conference on
  Autonomous Agents and MultiAgent Systems}, 25--33. International Foundation
  for Autonomous Agents and Multiagent Systems.

\bibitem[{Rusu et~al.(2016{\natexlab{a}})Rusu, Rabinowitz, Desjardins, Soyer,
  Kirkpatrick, Kavukcuoglu, Pascanu, and Hadsell}]{progressive1}
Rusu, A.A., Rabinowitz, N.C., Desjardins, G., Soyer, H., Kirkpatrick, J.,
  Kavukcuoglu, K., Pascanu, R., and Hadsell, R. (2016{\natexlab{a}}).
\newblock Progressive neural networks.
\newblock \emph{arXiv preprint arXiv:1606.04671}.

\bibitem[{Rusu et~al.(2016{\natexlab{b}})Rusu, Vecerik, Roth{\"o}rl, Heess,
  Pascanu, and Hadsell}]{progressive2}
Rusu, A.A., Vecerik, M., Roth{\"o}rl, T., Heess, N., Pascanu, R., and Hadsell,
  R. (2016{\natexlab{b}}).
\newblock Sim-to-real robot learning from pixels with progressive nets.
\newblock \emph{arXiv preprint arXiv:1610.04286}.

\bibitem[{Schulman(2015)}]{gae}
Schulman, J. (2015).
\newblock High-dimensional continuous control using generalized advantage
  estimation.
\newblock \emph{arXiv preprint arXiv:1506.02438.}

\bibitem[{Schulman et~al.(2017)Schulman, Wolski, Dhariwal, Radford, and
  Klimov}]{ppo}
Schulman, J., Wolski, F., Dhariwal, P., Radford, A., and Klimov, O. (2017).
\newblock Proximal policy optimization algorithms.
\newblock \emph{arXiv preprint arXiv:1707.06347}.

\bibitem[{Tang et~al.(2019)Tang, Xu, and Tomizuka}]{tang2019disturbance}
Tang, C., Xu, Z., and Tomizuka, M. (2019).
\newblock Disturbance-observer-based tracking controller for neural network
  driving policy transfer.
\newblock \emph{IEEE Transactions on Intelligent Transportation Systems}.

\bibitem[{Taylor and Stone(2009)}]{reinforcement_transfer}
Taylor, M.E. and Stone, P. (2009).
\newblock Transfer learning for reinforcement learning domains: A survey.
\newblock \emph{Journal of Machine Learning Research}, 10(Jul), 1633--1685.

\bibitem[{Todorov et~al.(2012)Todorov, Erez, and Tassa}]{mujoco}
Todorov, E., Erez, T., and Tassa, Y. (2012).
\newblock Mujoco: A physics engine for model-based control.
\newblock In \emph{2012 IEEE/RSJ International Conference on Intelligent Robots
  and Systems}, 5026--5033. IEEE.

\bibitem[{Xu et~al.(2019)Xu, Chang, Tang, Liu, and Tomizuka}]{xu2019toward}
Xu, Z., Chang, H., Tang, C., Liu, C., and Tomizuka, M. (2019).
\newblock Toward modularization of neural network autonomous driving policy
  using parallel attribute networks.
\newblock In \emph{2019 IEEE Intelligent Vehicles Symposium (IV)}, 1400--1407.
  IEEE.

\bibitem[{Xu et~al.(2018)Xu, Tang, and Tomizuka}]{xu2018zero}
Xu, Z., Tang, C., and Tomizuka, M. (2018).
\newblock Zero-shot deep reinforcement learning driving policy transfer for
  autonomous vehicles based on robust control.
\newblock In \emph{2018 21st International Conference on Intelligent
  Transportation Systems (ITSC)}, 2865--2871. IEEE.

\end{thebibliography}
\end{document}